\documentclass[letterpaper, 10 pt, journal, twoside]{IEEEtran}
\usepackage{amsmath,amsfonts}
\usepackage{algorithmic}
\usepackage{algorithm}
\usepackage{array}
\usepackage[caption=true,font=footnotesize,labelfont=rm,textfont=rm]{subfig}
\usepackage{textcomp}
\usepackage{stfloats}
\usepackage{url}
\usepackage{verbatim}
\usepackage{graphicx}
\usepackage{cite}
\usepackage{ulem}

\usepackage{bbding}
\usepackage{bm}
\usepackage{multirow}
\usepackage[Symbol]{upgreek}
\usepackage{amsfonts}
\usepackage{diagbox}
\usepackage{amsmath}

\usepackage{caption}
\usepackage{subcaption}
\usepackage{pifont}

\begin{document}
%
\title{KN-LIO: Geometric Kinematics and Neural Field Coupled LiDAR-Inertial Odometry}
\author{Zhong~Wang$^{1}$, Lele~Ren$^{2}$, Yue~Wen$^{1}$, and Hesheng~Wang$^{1}$
\thanks{This work was supported in part by the National Natural Science Foundation of China under Grant 62403308; in part by the Postdoctoral Fellowship Program of CPSF under Grant Number GZC20241013 (\textit{Corresponding author: Hesheng Wang.})}
\thanks{$^{1}$Zhong Wang, Yue Wen, and Hesheng Wang are with the Department of Automation, School of Electronics, Information and Electrical Engineering, Shanghai Jiao Tong University, Shanghai 200240, China (email: \{cszhongwang, wenyue023, wanghesheng\}@sjtu.edu.cn).}
\thanks{$^{2}$Lele Ren is with the School of Information and Control Engineering, China University of Mining and Technology, Xuzhou 221008, China (e-mail:  \tt\footnotesize{leo\_2023@cumt.edu.cn}).}
}
%
%

\markboth{Manuscript Submitted to arXiv Transactions on ***.}
{Wang \MakeLowercase{\textit{et al.}}: KN-LIO: Geometric Kinematics and Neural Field Coupled LiDAR-Inertial Odometry} 

%



\maketitle

\begin{abstract}
Recent advancements in LiDAR-Inertial Odometry (LIO) have boosted a large amount of applications. However, traditional LIO systems tend to focus more on localization rather than mapping, with maps consisting mostly of sparse geometric elements, which is not ideal for downstream tasks. Recent emerging neural field technology has great potential in dense mapping, but pure LiDAR mapping is difficult to work on high-dynamic vehicles. To mitigate this challenge, we present a new solution that tightly couples geometric kinematics with neural fields to enhance simultaneous state estimation and dense mapping capabilities. We propose both semi-coupled and tightly coupled Kinematic-Neural LIO (KN-LIO) systems that leverage online SDF decoding and iterated error-state Kalman filtering to fuse laser and inertial data. Our KN-LIO minimizes information loss and improves accuracy in state estimation, while also accommodating asynchronous multi-LiDAR inputs. Evaluations on diverse high-dynamic datasets demonstrate that our KN-LIO achieves performance on par with or superior to existing state-of-the-art solutions in pose estimation and offers improved dense mapping accuracy over pure LiDAR-based methods. The relevant code and datasets will be made available at https://**.
\end{abstract}

\begin{IEEEkeywords}
LiDAR-Inertial Odometry, Dense Mapping, Neural Field, SLAM
\end{IEEEkeywords}

%
\IEEEpeerreviewmaketitle

\section{Introduction}

\IEEEPARstart{R}{obotics} is gradually becoming integrated into daily human life. Their interaction with the environment and humans relies on a comprehensive perception of both intro-state and extro-state, which is facilitated by SLAM (Simultaneous Localization and Mapping) technologies. Recent rapid advancements in such techniques based on the fusion of LiDAR (Light Detection and Ranging) and IMU (Inertial Measurement Unit), namely the LiDAR-Inertial Odometry (LIO), have significantly propelled the implementation of related applications.

Early LIO systems focused on a loosely coupled integration of these two information sources~\cite{carto}. 
While this way is straightforward, it does not fully leverage the complementary information between them. 
Recent research on LIO systems indicates that a tightly coupled fusion is the key to maximizing the complementary characteristics of the two types of sensors~\cite{qin2020lins,lio-sam,dliom,dlio,fast-lio2}. 
On one hand, real-time estimation of IMU biases enhances the accuracy of state prediction and helps eliminate motion distortion in point clouds. 
On the other hand, the registration of the undistorted point clouds with the environmental map can effectively suppress system state drift, enabling the system to work stably through the precise collaboration of LiDAR and IMU. 
However, existing LIO solutions tend to prioritize localization over mapping. In other words, their map representations often consist of geometric elements that facilitate pose estimation, such as points, lines, surfels, or voxels. 
Yet, such geometric elements are not what downstream tasks like navigation, augmented reality, and virtual reality aspire to. 
In practice, such applications prefer a dense three-dimensional scene representation rather than sparse geometric elements.

Traditional dense reconstruction techniques can be categorized into offline reconstruction based on full maps and online incremental reconstruction. 
The former includes methods such as Poisson reconstruction~\cite{poissonrecon,puma}, which aims to fit surface equations, and graph-cut-based surface reconstruction~\cite{4408892}, which focuses on identifying segmentation surfaces. Online reconstruction methods first emerged in RGBD-SLAM~\cite{6162880}, where they partition the scene into regular grids and establish a distance field from spatial points to the nearest surfaces, employing the Marching Cubes algorithm for dense scene reconstruction~\cite{marchingcube}. 
In addition to the aforementioned traditional methods, dense reconstruction techniques for point clouds have recently achieved promising results driven by advancements in machine learning, especially with the rise of neural radiance fields (NeRF)~\cite{nerf}. 
In offline reconstruction, some approaches represent the scene as a compact octree and learn features attached to the vertices of spatial grids to fit dense neural fields~\cite{shinemapping,nfatlas}. For online reconstruction, NeRF-LOAM~\cite{nerfloam}, which is based on octree grid representation, and PIN-SLAM~\cite{pinslam}, which utilizes sparse neural point representation, both learn SDF (Signed Distance Function) decoders from scene features, track LiDAR poses, and jointly learn map neural features on the fly. 
Compared to traditional methods, the online solutions based on neural field representations not only enjoy the advantage of constructing continuous fields at arbitrary resolutions but also more naturally leverage data-driven approaches to fully integrate both introspective and extrospective perception information, thereby producing more accurate scene maps.

However, despite the commendable performance of such pure LiDAR-focused neural dense reconstruction techniques in relatively stable near-2D autonomous driving scenarios, their effectiveness in more challenging cases, such as handheld devices, backpacks, or fast 3D drones, remains unsatisfactory. 
A natural idea is to enhance online reconstruction under such complex motion conditions by integrating the instantaneous dynamic perception capability of IMUs. 
However, extensive research on traditional LIO has shown that loosely coupled approaches do not significantly improve system performance. 
The challenge arises from how to tightly integrate geometric kinematics with neural fields. 
On one hand, neural fields are encoded by high-dimensional neural features, while IMU-based geometric kinematics evolves on a composite manifold; on the other hand, IMU-based geometric kinematics have explicit mathematical expressions, whereas neural fields are implicitly represented within decoders and neural features.
To our knowledge, this paper will be the first to present the solutions for these issues.

To tackle the aforementioned challenges, we represent the scene map as a neural point cloud and obtain the distance to the nearest surface through online learning of an SDF decoder and neural features, while fusing laser and inertial observations based on an error-state Kalman filter. To effectively estimate the system state, we propose both semi-coupled and tightly coupled \textbf{L}iDAR-\textbf{I}nertial \textbf{O}dometry systems that integrate geometric \textbf{K}inematics with \textbf{N}eural fields, semi-KN-LIO and \textbf{KN-LIO} for short, respectively. 
In the semi-coupled approach, we first predict the system state (position, velocity, orientation, IMU biases) using high-frequency IMU readings, then undistort the input point cloud and register it to the built map based on this prediction, and finally update the system error state using the registered pose as a composite observation. This semi-coupled method allows for a relative decoupling of traditional kinematics and neural fields while utilizing the neural field observations to correct the system's error state. 
In the tightly coupled approach, we further integrate the error state updates with the point cloud registration process, obtaining the Kalman gain from the current point cloud and neural field in each iteration to update the system's error state. 
Although this tightly coupled method is more complex, it reduces information transfer loss and effectively enhances the accuracy of state estimation. 
Additionally, thanks to the point cloud registration based on neural point representation, which does not require establishing point-to-point (feature-to-feature) correspondences, we extend the system to accommodate asynchronous multi-LiDAR inputs (whether homogeneous or heterogeneous), effectively leveraging complementary viewpoints to construct a more complete map and improve state estimation accuracy.

We evaluated our systems on a large number of datasets gathered from high-dynamic platforms, including handheld devices, mobile robotics, and drones. The results demonstrate that, in terms of pose estimation, KN-LIO achieves performance comparable to existing state-of-the-art solutions. In terms of dense mapping, KN-LIO yields more accurate results than pure LiDAR mapping solutions. To facilitate the robotics community, we will release the relevant codes and data upon the acceptance of this paper.

\begin{figure*}[t]
        \centering
        \includegraphics[width=\textwidth]{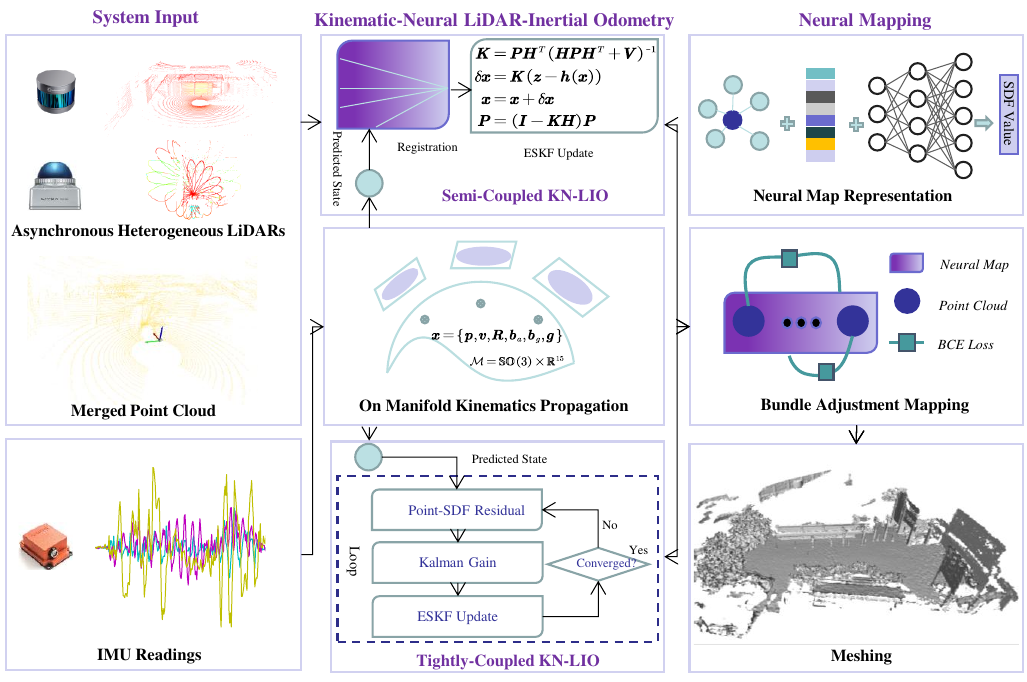}
        \caption{Overview of Kinematic-Neural LiDAR-Inertial Odometry. The high-frequency IMU readings are used for recursive system state propagation on the manifold. Asynchronous heterogeneous point clouds are first integrated based on timestamps and extrinsics, and then undistorted according to the recursive state. In a semi-coupled mode, the undistorted point cloud is registered with a neural field by minimizing point-SDF residuals, and the error-state Kalman filter is updated with the registration result. In a tightly coupled mode, the system calculates the point-SDF residuals and updates the error-state Kalman filter based on them, iterating until convergence. The system represents the map as a set of neural points, where the feature of a given spatial point is obtained by weighted distance to its neighboring neural points, and its corresponding SDF value is decoded from its features through a simple MLP. Under the tracked state, the system performs bundle optimization on the local neural map together with the current point cloud. Finally, the dense mesh of the scene is reconstructed through SDF queries and marching cubes.}
        \label{fig-system}
        \vspace{-3mm}
\end{figure*}

\section{Related Work}\label{sec-related-work}
\subsection{LiDAR-Inertial Odometry}
In recent years, the effective fusion of LiDAR and IMU has significantly improved the performance of LiDAR-Inertial Odometry/SLAM systems for localization and mapping. Existing approaches can be divided into feature-based ones and direct ones depending on whether geometric features are extracted. Feature-based approaches often rely on LOAM~\cite{loam} features. 
Initially, these methods drew inspiration from valuable experiences in visual-inertial SLAM. For example, based on VINS-Mono~\cite{vins}, LIO-Mapping~\cite{liom} utilizes graph optimization to construct geometric constraints using the line and plane features defined by LOAM~\cite{loam}, making it the first tightly coupled LiDAR-Inertial odometry solution. Subsequently, Qin \textit{et al.} proposed LINS~\cite{qin2020lins}, which achieves effectively tightly coupling of feature-based ICP registration~\cite{icp} and Kalman filter~\cite{quat_err_state}. Shan \textit{et al.} introduced LIO-SAM~\cite{lio-sam}, which abandons inter-frame registration, aligns point clouds to local maps using LeGoLOAM~\cite{lego-loam} features, and jointly registers pose and IMU pre-integration based on smooth and mapping techniques~\cite{isam,isam2} to optimize system state. 
The reasonable utilization of line/plane features and effective estimation of IMU biases enabled LIO-SAM to achieve impressive results at the time, but its design requires high performance from the IMU and performs moderately in complex and unstructured scenarios. 
Additionally, CLINS~\cite{clins} fuses LiDAR-Inertial data in a continuous-time framework, transforming the state estimation problem into a joint optimization problem of continuous-time trajectories and observed features, effectively fusing information from multiple frames but with higher computational complexity.

In the domain of direct methods, early works were based on probabilistic grid maps, loosely integrating LiDAR and IMU information~\cite{carto}. Building upon this representation, Wang \textit{et al.} drew inspiration from LIO-SAM and proposed the tightly coupled direct approach D-LIOM~\cite{dliom}. Furthermore, to combine laser-visual-inertial information, Wang \textit{et al.} introduced Ct-LVI~\cite{ctlvi} within a continuous-time SLAM framework, constructing a tightly coupled LiDAR-Inertial subsystem Ct-LIO through direct registration. 
In a similar way, Chen \textit{et al.}~\cite{dlio} improved upon the direct  LiDAR odometry system (DLO) by tightly coupling state estimation based on GICP in a coarse-to-fine manner, achieving satisfactory performance in various complex scenarios. 
Building on their earlier work Fast-LIO~\cite{fastlio}, Xu \textit{et al.} no longer extract point cloud line features, but directly construct point-to-plane constraints by fitting surfaces to neighboring points queried in an ikdtree, proposing Fast-LIO2~\cite{fast-lio2}, which has achieved state-of-the-art performance in terms of localization accuracy and robustness.

\subsection{Dense Mapping}

Traditional dense mapping methods can be classified into Poisson-based, tetrahedron-based, voxel-based, and facet-based. 
Poisson-based reconstruction~\cite{poissonrecon,puma} represents the scene surface as an indicator function and fits the target object/scene surface gradients (in the normal vectors of the point cloud) through optimization techniques to reconstruct the surface. The surface is then extracted from this function using techniques similar to the Marching Cube alogrithm~\cite{marchingcube}. This implicit representation gives the reconstructed scene continuous, compact, and excellent characteristics. 
Tetrahedron-based methods~\cite{4408892} first decompose the point cloud into tetrahedra, then perform binary classification on these decomposed tetrahedra, and finally extract the surface from their labels. 
Voxel-based methods~\cite{6162880} first appeared in online reconstruction from RGB-D data, aiming to accelerate the mapping efficiency using GPU-accelerated regularized spatial division. 
This technology estimates the truncation distance from the observation point cloud to the nearest surface of the divided grid points and extracts the surface from the grid using Marching Cube~\cite{marchingcube}. Although GPU acceleration can achieve real-time performance in indoor scenes, challenges in storage space and long-term mapping consistency still exist when effectively extending it to large outdoor scenes~\cite{101145,25083632508374,7368096,8255617}. 
Additionally, some methods represent the scene as a series of facets/points and reconstruct the surface through sampling and point rendering techniques~\cite{7807268,7266816}. This dispersed representation approach is more flexible than representing the entire function or regular grid, but reconstructing high-precision surfaces requires more elements, inevitably increasing computational and storage burdens. Recently, LIN et al.~\cite{imesh} proposed an efficient facet-based mesh reconstruction algorithm, achieving real-time online mesh reconstruction.

The significant advancements in machine learning techniques have also promoted learning-based surface reconstruction, especially with the emergence of neural radiance fields in recent years. To achieve novel view synthesis, Mildenhall \textit{et al.}~\cite{nerf} represent the scene as an implicit function with position and orientation as input parameters, estimating the view color through ray tracing. 
Similarly, the depth of spatial points can also be obtained from the opacity in the scene function through ray tracing. 
It can be said that this implicit representation of the scene aligns well with Poisson reconstruction. 
Using such representation, a series of works have made active attempts at online SLAM based on NeRF~\cite{niceslam,10550721}, but the low rendering efficiency has always been a bottleneck. 
To accelerate the rendering procedures, some methods divide space into grids and characterize the scene with features attached to grid vertices, decoding target values with an MLP~\cite{voxfusion,shinemapping,nfatlas}. 
There are also approaches that represent the scene as neural point clouds~\cite{pointnerf,pointslam,3dgs,monogs,splatam}. These grid and point representations were initially focused on object-level or room-level reconstruction. 
In order to extend them to outdoor scenes, many works have made active explorations, such as structuring sparse grids based on octrees~\cite{nerfloam} or organizing neural points based on hash grids~\cite{pointslam}. It is evident that these approaches cleverly draw on the development path of traditional geometric methods while effectively empowered by the advantages of data-driven learning.

\section{Method}\label{sec-method}
\subsection{Overview}
The system workflow is illustrated in Fig. \ref{fig-system}. We represent the scene as a set of learnable neural points, where the SDF value at each spatial position is encoded by the weighted distance from its surrounding neural points' features and estimated by a simple MLP decoder. 
For the asynchronous heterogeneous LiDAR input, we first generate an integrated point cloud based on spatiotemporal parameters. 
For the high-frequency IMU data, we recursively estimate the nominal state and error state, as well as the corresponding error state covariance, using an error-state Kalman filter, and remove point cloud distortions based on the propagated state. 
Subsequently, within the semi-coupled KN-LIO, we register the de-skewed point cloud to the current neural map by minimizing the SDF residuals of sampled points to obtain the registration pose, and then update the ESKF with this pose. In the tightly coupled KN-LIO, starting from the nominal state, we iteratively update the Kalman filter by computing the SDF residuals of sampled points, until the ESKF converges. Finally, based on the optimized pose, we jointly optimize and update the local neural map by minimizing the BCE loss between sampled points and the neural field, and generate a dense mesh from the estimated neural field resorting to marching cubes.

\subsection{Notation}
We define the state of KN-LIO as:
\begin{equation}
    \bm{x}=\{\bm{p},\bm{v},\bm{R},\bm{b}_g,\bm{b}_a,\bm{g}\}\in\mathcal{M}=\mathbb{SO}(3)\times\mathbb{R}^{15}
    ,
\end{equation}
where the elements the system's position ($\bm{p}$), velocity ($\bm{v}$), gyroscope bias ($\bm{b}_g$), accelerometer bias ($\bm{b}_a$), and gravity $\bm{g}$ are in $\mathbb{R}^3$, while rotation ($\bm{R}$) is in $\mathbb{SO}(3)$. $\bm{p}$, $\bm{v}$, $\bm{R}$, and $\bm{g}$ are in the world frame by default, while $\bm{b}_g$ and $\bm{b}_a$ are in the body frame (IMU frame).

We define the corresponding error state to be estimated as:
\begin{equation}
    \delta\bm{x}=\begin{bmatrix}
        \delta\bm{p}^T,\delta\bm{v}^T,\delta\bm{\theta}^T,\delta\bm{b}_g^T,\delta\bm{b}_a^T,\delta\bm{g}^T
    \end{bmatrix}\in\mathbb{R}^{18},
\end{equation}
where $\delta\bm{\theta}$ is the axis-angle of the rotation error state $\delta\bm{R}$, which can be transformed from/to $\delta\bm{R}$ by logarithmic/exponential mapping.

\subsection{Error-State Kalman Filter}
The Kalman filter for error states considers the true state of the system as the sum of the nominal state and the error state. It recursively calculates the covariance of the nominal state, error state, and error state through system inputs, and corrects the error state through external sensor observations to update the system state. Firstly, we briefly review the key concepts of the error-state Kalman filter. We refer the audience to Joan's~\cite{quat_err_state} and Gao's work~\cite{Gao2023SAD}.

\subsubsection{Continuous-time Kinematics Model}
In continuous time, the motion of a rigid body system in an inertial frame follows Newton's laws of motion. The nominal kinematic model is:
\begin{align}
 \dot{\bm{p}}&=\bm{v},\notag\\
 \dot{\bm{v}}&=-\bm{R}(\bm{\tilde{a}}-\bm{b}_a-\bm{\eta}_a)+\bm{g},\notag\\
 \dot{\bm{R}}&=\bm{R}(\tilde{\bm{\omega}}-\bm{b}_g-\bm{\eta}_g)^\wedge,\notag\\
 \dot{\bm{b}_g}&=\bm{\eta}_{bg},\notag\\
 \dot{\bm{b}_a}&=\bm{\eta}_{ba},\notag\\
 \dot{\bm{g}}&=\bm{0},
\end{align}
where $\bm{\tilde{a}}$ and $\tilde{\bm{\omega}}$ denote the IMU readings of acceleration and angular velocity, $\bm{\eta}_{ba}$ ($\bm{\eta}_{bg}$) is the random inpulse of accelerometer (gyroscope), $(\cdot)^\wedge$ means the skew-symmetric matrix of the given vector. 
The corresponding error-state kinematics model is formulated as:
\begin{align}
 \delta\dot{\bm{p}}&=\delta{\bm{v}},\notag\\
 \delta\dot{\bm{v}}&=-\bm{R}(\bm{\tilde{a}}-\bm{b}_a)^{\wedge}\delta\bm{\theta}-\bm{R}\delta\bm{b}_a-\bm{\eta}_a+\delta{\bm{g}},\notag\\
 \delta{\dot{\bm{\theta}}}&=-(\tilde{\bm{\omega}}-\bm{b}_g)^{\wedge}\delta\bm{\theta}
-\delta\bm{b}_g-\bm{\eta}_g,\notag\\
 \delta\dot{\bm{b}_g}&=\bm{\eta}_{bg},\notag\\
 \delta\dot{\bm{b}_a}&=\bm{\eta}_{ba},\notag\\
 \delta\dot{\bm{g}}&=\bm{0}.
\end{align}

\subsubsection{Discrete-time Kinematics Model}
By performing kinematic integration on the time dimension of the above equation, the nominal state kinematic equation in discrete time can be obtained as follows:
\begin{align}
 \bm{p}(t+\Delta{t})&=\bm{p}(t)+\bm{v}\Delta{t}+\frac{1}{2}(\bm{R}(\tilde{\bm{a}}-\bm{b}_a)\Delta{t}^2+\frac{1}{2}\bm{g}\Delta{t}^2,\notag\\
 \bm{v}(t+\Delta{t})&=\bm{v}(t)+\bm{R}(\tilde{\bm{a}}-\bm{b}_a)\Delta{t}+\bm{g}\Delta{t},\notag\\
 \bm{R}(t+\Delta{t})&=\bm{R}(t)\text{Exp}((\tilde{\bm{\omega}}-\bm{b}_g)\Delta{t}),\notag\\
 \bm{b}_g(t+\Delta{t})&=\bm{b}_g(t),\notag\\
 \bm{b}_a(t+\Delta{t})&=\bm{b}_a(t),\notag\\
 \bm{g}(t+\Delta{t})&=\bm{g}(t),
\end{align}
where $\Delta t$ is the integration interval and $\text{Exp}(\cdot)$ denotes the compound exponential map from $\mathbb{R}^3$ to $\mathbb{SO}(3)$.
Correspondingly, the error state kinematic equation for discrete-time is:
\begin{align}
 \delta\bm{p}(t+\Delta{t})&=\delta\bm{p}+\delta\bm{v}\Delta{t},\notag\\
 \delta\bm{v}(t+\Delta{t})&=\delta\bm{v}(t)+(-\bm{R}(\tilde{\bm{a}}-\bm{b}_a)^\wedge\delta\bm{\theta}-\bm{R}\delta{\bm{b}_a}+\delta{g})\Delta{t}-\bm\eta_v,\notag\\
 \delta\bm{R}(t+\Delta{t})&=\bm{R}(t)\text{Exp}((\tilde{\bm{\omega}}-\bm{b}_g)\Delta{t}),\notag\\
 \delta\bm{b}_g(t+\Delta{t})&=\delta\bm{b}_g+\bm\eta_g,\notag\\
 \delta\bm{b}_a(t+\Delta{t})&=\delta\bm{b}_a+\bm\eta_a,\notag\\
 \delta\bm{g}(t+\Delta{t})&=\delta\bm{g}.\label{eq-err-state-kinematics}
\end{align}
\subsubsection{Propagation, Update, and Reset}
According to the traditional form of the extended Kalman filter, the error state transition equation in matrix form can be obtained by reorganizing the above equations:
\begin{align}
    \delta\bm{x}&\leftarrow\bm{F}\delta\bm{x}+\bm{\omega},\bm{\omega}\sim \mathcal{N}(\bm{0},\bm{Q}),\label{eq-dx}\\
    \bm{Q}&=\Lambda(\bm{O},\bm{Q}_{\bm\eta_v},\bm{Q}_{\bm\eta_\theta},\bm{Q}_{\bm\eta_g},\bm{Q}_{\bm\eta_a},\bm{O}),  
\end{align}
where $\bm{F}$ is the transition matrix of the error state which can be obtained from Eq.~\ref{eq-err-state-kinematics}, $\bm{Q}$ is the covariance matrix of the error-state noise $\bm{\omega}$, and $\Lambda$ denotes the diagonal matrix. 
For each step of error state propagation, the covariance of the error state ($\bm{P}$) needs to be mapped to a new tangent space by the transition matrix $\bm{F}$ and incorporate new noise:
\begin{equation}
    \bm{P}\leftarrow\bm{F}\bm{P}\bm{F}^T+\bm{Q}\label{eq-Q}.
\end{equation}

If there are no new observations except the IMU readings, the uncertainty of the system will continue to increase with the recurrence of covariance. When a new sensor observes input, we can reduce this uncertainty by correcting the error state. The update at this time can be carried out according to the standard form of the extended Kalman filter:
\begin{align}
    \bm{K}&={\bm{P}}\bm{H}^T(\bm{H}{\bm{P}}\bm{H}^T+\bm{V})^{-1},\label{eq-update-1}\\
    \delta\bm{x}&=\bm{K}(\bm{z}-\bm{h}({\bm{x}})),\label{eq-update-2}\\
    \bm{x}&\leftarrow{\bm{x}}+\delta\bm{x},\\
    \bm{P}&\leftarrow(\bm{I}-\bm{K}\bm{H}){\bm{P}}\label{eq-update-4},
\end{align}
where $\bm{H}$ is the Jacobian matrix of the measurement $\bm{h}(\cdot)$ with respect to the error-state $\delta\bm{x}$, which can be obtained by the chain rule:
\begin{equation}
    \bm{H}=\frac{\partial\bm{h}}{\partial\bm{x}}\frac{\partial\bm{x}}{\partial\delta\bm{x}},\label{eq-H}
\end{equation}
where the first term depends on the specific observation equation which will be discussed in subsequent sections, while the second term is known:
\begin{align}
    \frac{\partial\bm{x}}{\partial\delta\bm{x}}&=\Lambda(\bm{I}_{3\times3},\bm{I}_{3\times3},\bm{J}_r^{-1}(\bm{R}),\bm{I}_{3\times3},\bm{I}_{3\times3},\bm{I}_{3\times3}),\label{eq-jac-hdx-0}\\
    \bm{J}_r^{-1}(\bm{R}) &= \bm{I}_{3\times3}+\frac{1}{2}\bm{\theta}^\wedge+(\frac{1}{\Vert\bm{\theta}\Vert^2}-\frac{1+\cos\Vert\bm{\theta}\Vert}{2\Vert\bm{\theta}\Vert\sin\Vert\bm{\theta}\Vert}){\bm{\theta}^\wedge}^2.\label{eq-jac-hdx-1}
\end{align}

After completing an update of the error state and covariance, we need to reset the error state to prepare for the next update. Among them, the reset of the error state is trivial, resetting all variables to the zero vector is sufficient, while the covariance needs to be reset to the new state reference point. That is, the following reset operation is performed:
\begin{align}
    \delta\bm{x} &= \bm{0},\\
    \bm{P}&=\bm{J}_r\bm{P}\bm{J}_r^T,\\
    \bm{J}_r&=\Lambda(\bm{I}_{6\times6},\bm{J}_{\bm\theta},\bm{I}_{9\times9}),\\
    \bm{J}_{\bm\theta}&=\bm{I}_{3\times3}-\frac{1}{2}\delta\bm\theta^\wedge.
\end{align}

\subsection{Neural Map Representation}
Currently, in neural SDF map representations for LiDAR, there are typically two organizational forms: octree-based and point-based. The former divides space into fixed-size voxels at a certain resolution, storing voxels hit by point clouds in an octree, with each vertex of the voxel attached with a feature for learning scene representation. 
Although this method significantly reduces storage consumption, ray casting based on octrees remains a major efficiency bottleneck. 
The latter represents space as a series of neural points, with each point having learnable features attached, and efficient queries can be achieved by organizing these points using a KD-tree. Therefore, inspired by the approach of Pan \textit{et al.} \cite{pinslam}, this paper adopts the latter map representation form. In this way, the features of each point in space are characterized by a weighted representation based on its neighboring points, and a simple MLP decoder is leveraged to predict the closest distance to the scene surface from these features.

\subsection{System Initialization}
Like most LiDAR-Inertial systems, we initialize the system in a stationary state. We define the robot's starting point as the origin of the world frame, with the world frame's z-axis perpendicular to the gravity level, and the direction of gravity aligned with the negative z-axis. With this definition, we can accumulate multiple frames of IMU readings to estimate the initial orientation of the system. 
Let the accelerometer and gyroscope readings ($\mathcal{A},\mathcal{W}$) during the initialization phase be:
\begin{align}
    \mathcal{A}&=\{\tilde{\bm{a}}_0,\dots,\tilde{\bm{a}}_n\},\\
     \mathcal{W}&=\{\tilde{\bm{\omega}}_0,\dots,\tilde{\bm{\omega}}_n\}.
\end{align}

We believe that the bias of the accelerometer is a very small amount relative to the measured acceleration value, so the initial orientation of the carrier can be obtained by aligning the direction of the acceleration mean vector with the opposite direction of gravity:
\begin{align}
    \bm{R}_{0}=\textit{R}(\bm{e}_{\overline{\mathcal{A}}},\bm{e}_{\bm{g}}),
\end{align}
where $\bm{e}$ is a unit vector and $\overline{\mathcal{A}}$ is the mean of $\mathcal{A}$.
On the basis of obtaining the initial rotation, if the local gravity acceleration value is known, we can also estimate the initial bias of the accelerometer:
\begin{align}
    {\bm{b}}_{a_0}=\overline{\bm{\mathcal{A}}}+\bm{R}_{0}^T\bm{g}.
\end{align}
As usual, the initial position and velocity are set as zero vectors, while the mean gyroscope reading is regarded as the initial guess of $\bm{b}_g$.

\subsection{Sensor Data Preprocessing}
\begin{figure}[t]
        \centering
        \includegraphics[scale = 1]{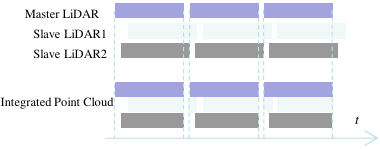}
        \caption{Asynchronous LiDAR merging.}
        \label{fig-multi-lidar}
\end{figure}
The rotating mechanical LiDAR has low point cloud resolution in the vertical direction, while the existing solid-state LiDAR can only present point clouds within the forward view. To enhance the robustness of LIO systems in various complex environments, integrating multi-view inputs is a more practical solution. However, this is not straightforward for feature-based approaches. 
Thanks to the direct registration of point clouds to neural maps, our system can be easily extended to accommodate multi-LiDAR inputs from complementary viewpoints. 
Specifically, for the input of multiple LiDAR point clouds (homogeneous or heterogeneous), we use the sampling time intervals of the main LiDAR as a reference and transform all points sampled by auxiliary LiDARs within the same time interval to the coordinate system of the main LiDAR based on the calibrated extrinsics, generating a new integrated point cloud together with the main LiDAR's point cloud. Fig. \ref{fig-multi-lidar} illustrates this operation.

For the high-frequency IMU data input, we recursively update the nominal state, error state, and covariance of the system as shown in Eq. \ref{eq-dx} to Eq. \ref{eq-Q}. Subsequently, with the predicted relative transformation, we correct the distortion of the integrated point cloud by interpolation measuring poses between the starting and ending timestamps of the point cloud.

\subsection{Semi-Coupled KN-LIO}
In the semi-coupled mode, we consider treating point cloud and IMU observations relatively independently. That is, we first utilize an IMU readings to propagate the error-state Kalman filter and then correct the filter using the registration results. The advantage of this approach is that it decouples the data processing for the two types of sensors through result-level fusion, while simultaneously achieving timely correction of the system state and IMU biases by propagating external LiDAR observation information. By linking the chain of state propagation prediction, point cloud distortion correction, point cloud registration with the neural map, and filter update, the system can achieve long-term stable running.

We perform direct registration from point clouds to neural maps to obtain LiDAR poses. Assuming the pose of point $\bm{p}$ is $\bm{T}=\{\bm{R},\bm{t}\}$, the purpose of registration is to adjust the pose so that the SDF value at its spatial position approaches 0, that is:
\begin{align}
    \{\bm{R},\bm{t}\}^*&=\mathop{\arg\min}\limits_{\bm{R},\bm{t}}\sum_{i}f(\bm{p}_i)^2\\
    &=\mathop{\arg\min}\limits_{\bm{R},\bm{t}}\sum_{i}D(\bm{R}\bm{p}_i+\bm{t})^2.\label{eq-registration-goal}
\end{align}
To use traditional optimization algorithms such as the Levenberg-Marquardt algorithm to solve the optimal pose, we need to obtain the derivatives of $f(\bm{p}_i)$ for rotation and translation parameters:
\begin{align}
    \frac{\partial{D(\bm{R}\bm{p}_i+\bm{t})}}{\partial\bm{t}}&=\frac{\partial{D(\bm{p}_i^\prime)}}{\partial\bm{p}_i^{\prime}}=\bm{n}_i^T,\\
    \frac{\partial{D(\bm{R}\bm{p}_i+\bm{t})}}{\partial\bm{\theta}}&=\frac{\partial{D(\bm{p}_i^\prime)}}{\partial\bm{p}_i^{\prime}}\frac{\partial{\bm{p}_i^{\prime}}}{\partial\bm\theta}=(-\bm{p}_i\times\bm{n}_i)^T,\\
    \bm{J}_{\bm{t},\bm{\theta}}^{f(\bm{p}_i)}&=[\bm{n}_i^T,(-\bm{p}_i\times\bm{n}_i)^T],
\end{align}
where $\bm{n}_i$ denotes the normal vector of the neural field at $\bm{p}_i$ and $\bm\theta$ is the axis-angle of the rotation.

After obtaining the registered pose, updating the filter will become simpler, and the observation model at this point becomes:
\begin{align}
    \bm{z}_s&=\bm{h}_s(\bm{x})+\bm{v}_s=\bm{x}+\bm{v}_s,~\bm{v}_s\in\mathcal{N}(\bm{0},\bm{V}_s),
\end{align}
where $\bm{v}_s$ denotes the noise of the point cloud to neural distance field registration results. 
Since the Jacobians of $\bm{h}_s(\bm{x})$ with respect to the nominal state at this time is the identity matrix, the update matrix in Eq.~\ref{eq-update-1} and Eq. \ref{eq-update-4} becomes the same as that of Eq.~\ref{eq-jac-hdx-0}.

\subsection{Tightly Coupled KN-LIO}
Although the semi-coupled mode can leverage LiDAR registration results to correct IMU biases and offers the advantage of relatively decoupling heterogeneous observations, the approach of first registering and then optimizing biases may entail certain risks of information loss. Therefore, this paper also tightly couples the registration of point clouds to the neural field with the estimation of IMU biases. With the goal of finding the optimal registration, we combine the estimation of the error-state Kalman filter with the iterative registration process, known as the iterative Kalman filter. 

In each iteration, we solve the following least squares problem:
\begin{equation}
\delta\bm{x}^*=\mathop{\arg\min}_{\delta\bm{x}}\Vert\bm{z}-\bm{H}(\bm{x}\oplus\delta\bm{x})\Vert^2_{\bm{V}} +\Vert\delta\bm{x}\Vert^2_{\bm{P}},
\end{equation}
where $\oplus$ indicates the plus operation on the manifold $\mathcal{M}$. 
In the Bayesian framework, this objective can be derived to yield analytical solutions as shown in Eq. \ref{eq-update-1}$\sim$Eq. \ref{eq-update-4}.
As shown in these equations, to perform tightly coupled iterations of the Kalman filter, we need to determine the observation equation at this time and the corresponding matrix $\bm{H}$. Our registration objective remains as shown in Eq. \ref{eq-registration-goal}, thus the observation model can be defined as:
\begin{align}
    \bm{z}_t&=\bm{h}_t(\bm{p}|\bm{x})+\bm{v}_t\\&=\begin{bmatrix}
        h(\bm{p}_0|\bm{x})\\
        \vdots\\
        h(\bm{p}_n|\bm{x})
    \end{bmatrix}=\begin{bmatrix}
        D_0(\bm{R}\bm{p}_0+\bm{t})\\
        \vdots\\
        D_n(\bm{R}\bm{p}_n+\bm{t})\\
    \end{bmatrix}.
\end{align}

To update the Kalman filter using this observation model, we need to calculate the partial derivative of $h(\cdot)$ with respect to ${\bm{x}}$:
\begin{align}
    \frac{\partial{h}(\bm{p}_i|\bm{x})}{\partial\bm{x}}&=\frac{\partial{D_i}}{\partial{\bm{p}^\prime}}\frac{\partial{\bm{p}^\prime}}{\partial{\bm{x}}}\\
    &=\bm{n}_i^T\begin{bmatrix}
        \bm{I}_{3\times3}&\bm{O}_{3\times3}&{-\bm{R}\bm{p}_i^\wedge} & \bm{O}_{3\times9}
    \end{bmatrix},
\end{align}
where $\bm{O}_{3\times3}$/$\bm{O}_{3\times9}$ is a $3\times3$/$3\times9$ zero matrix.
Thus, the $i$-th row of the matrix $\bm{H}$ is determined by the chain rule combining Eq.~\ref{eq-H} to Eq.~\ref{eq-jac-hdx-1}:
\begin{align}
        \bm{H}_i=\frac{\partial\bm{h}_i}{\partial\bm{x}}\frac{\partial\bm{x}}{\partial\delta\bm{x}}=\begin{bmatrix}
        \bm{n}_i^T&\bm{0}_{3}^T&\bm{n}_i^T{\bm{R}\bm{p}_i^\wedge}\bm{J}_r^{-1}(\bm{R}) & \bm{0}_{9}^T\end{bmatrix},
\end{align}
where $\bm{0}_3$/$\bm{0}_9$ denotes a zero vector with 3/9 elements.

We can estimate the error state and update the covariance according to Eq. \ref{eq-update-1}$\sim$Eq. \ref{eq-update-4}, however, a single point cloud typically contains thousands of points, which would make the inversion part in Eq. \ref{eq-update-1} computationally expensive. To address this issue, we resort to the Sherman-Morrison-Woodbury identity transformation and calculate the matrix $\bm{K}$ in an equivalent way as follows:
\begin{align}
    \bm{K}=({\bm{P}}^{-1}+{\bm{H}}^T{\bm{V}}^{-1}\bm{H})^{-1}{\bm{H}}^T{\bm{V}}^{-1}\label{eq-same-K}.
\end{align}

As a result, the dimension of the inversion in each iteration is reduced from a variable $N$ (number of points) to a constant 18 (the dimension of error state), making the system more efficient and stable.
In each iteration, we calculate the error state according to Eq. \ref{eq-same-K} and Eq. \ref{eq-update-2} and update the state by Eq. \ref{eq-update-4}. If the value of the current error state is below a certain threshold or reaches the maximum number of iterations, we consider the iteration to be complete, and transform the covariance matrix to the position of the last iteration using Eq. \ref{eq-update-4}.

\section{Experiments}\label{sec-experiment}
\subsection{Setup}
\subsubsection{Dataset}
We evaluated our KN-LIO and semi-KN-LIO on three widely-used datasets. 
One of them is the VIRAL dataset collected from a drone, consisting of 18 sequences from various indoor and outdoor courtyard scenes, including both structured and unstructured environments. Among these, the first 9 sequences exhibit relatively smooth motion, while the subsequent 9 sequences involve more aggressive drone movements, such as rapid ascents and fast rotations. 
Most existing methods have only been tested on the former 9 sequences. In this paper, to conduct a more comprehensive evaluation, we experiment with relevant competing approaches on all sequences and report detailed results accordingly.

One dataset is HILTI2022 collected from a construction site, where each sequence comprises featureless and varying lighting environments, posing challenges for an LIO system. This dataset is captured using a handheld device equipped with a Hesai PandarXT-32 LiDAR and Sevensense Alphasense Core. We evaluated the relevant algorithms on sequences with publicly available 6-DoF millimeter-level ground truth poses, specifically including ``Exp04 Construction Upper Level 1'' (Exp04), ``Exp05 Construction Upper Level 2'' (Exp05), ``Exp06 Construction Upper Level 3'' (Exp06), ``Exp14 Basement 2'' (Exp14), ``Exp16 Attic to Upper Gallery 2'' (Exp16), and ``Exp18 Corridor Lower Gallery 2'' (Exp18).

We conducted quantitative testing of the algorithm for mapping on Newer College. The dataset provides scenes reconstructed using survey-grade laser scanners. Typically, pure LiDAR mapping methods are tested on segments extracted from the sequence ``02\_long\_experiment''. For comparison purposes, we also extracted the corresponding IMU data from the raw dataset, while retaining IMU readings corresponding to 10 frames of point clouds for system initialization. It should be clarified that these 10 frames of point clouds within the corresponding time period were not involved in map reconstruction.

\subsubsection{Metrics}
Regarding localization, following the common practice in related methods~\cite{pinslam,fast-lio2}, we evaluated the absolute positioning error of the approaches. For each estimated pose sequence of a method, we utilized the Umeyama algorithm \cite{88573} to estimate its transformation relative to the ground truth, and subsequently calculated its Root Mean Square Error (RMSE). In terms of mapping, following the method proposed in \cite{30729593073599}, we assessed the accuracy (Acc.), completeness (Comp.), chamfer distance (C-L1), and F-Score of the reconstructed maps generated by the algorithm. Lower values for the first three metrics indicate better reconstruction quality, while a higher value for the last metric signifies a higher recall rate and precision.

\subsection{Traits Comparison}
\begin{table}[]
    \caption{Traits of LiDAR/LiDAR-Inertial Odometry systems.}
    \label{tab:traits}
    \centering
    \begin{tabular}{c|c|c|c|c|c}
     \hline
     \hline
         Method & Dense & Direct & Inertial & Neural &M-LiDAR\\
         \hline
         PIN-SLAM & \ding{51} & \ding{51} & \ding{55} & \ding{51} & \ding{55}\\
         SLAMesh & \ding{51} & \ding{51} & \ding{55} & \ding{55} & \ding{55}\\
         D-LIOM & \ding{55} & \ding{51} & \ding{51} & \ding{55} & \ding{51}\\
         Ct-LIO & \ding{55} & \ding{51} & \ding{51} & \ding{55} & \ding{51}\\
         dlio & \ding{55} & \ding{51} & \ding{51} & \ding{55} &\ding{55}\\
         LIO-SAM & \ding{55} & \ding{55} & \ding{51} & \ding{55} & \ding{55}\\
         Fast-LIO2 & \ding{55} & \ding{51} & \ding{51} & \ding{55} & \ding{55}\\
         Semi-KN-LIO (ours) & \ding{51} & \ding{51} & \ding{51} & \ding{51} & \ding{51}\\
         KN-LIO (ours) & \ding{51} & \ding{51} & \ding{51} & \ding{51} & \ding{51}\\
     \hline
     \hline
    \end{tabular}
\end{table}
Firstly, we compare the characteristics of our approach with competing methods. In terms of map representation, apart from our KN-LIO and semi-KN-LIO, only PIN-SLAM and SLAMesh support incremental dense reconstruction, with only PIN-SLAM and our approach being based on neural field representation, enabling continuous Signed Distance Function (SDF) field estimation. 
Regarding pose estimation, while several methods are feature-free, except for our approach, only D-LIOM and Ct-LIO support multiple LiDAR inputs. Furthermore, among all methods, our approach simultaneously supports dense reconstruction, direct registration, inertial information fusion, neural representation, and multi-LiDAR information fusion.

\subsection{Results of Localization}
\begin{table*}[]
    \caption{RMSEs (m) of advanced LiDAR/LiDAR-Inertial Odometry systems on VIRAL and HILTI22. \ding{55} means failure. `-' implies the result is not available.}
    \label{tab:localization}
    \centering
    \begin{tabular}{c|c|c|c|c|c|c|c|c|c}
     \hline
     \hline
         ~&PIN-SLAM\cite{pinslam}&SLAMesh\cite{slamesh}&D-LIOM\cite{dliom}&Ct-LIO\cite{ctlvi}&dlio\cite{dlio}&LIO-SAM\cite{lio-sam}&Fast-LIO2\cite{fast-lio2}&Semi-KN-LIO&KN-LIO \\
         \hline
         eee\_01&\ding{55}&0.089&0.191&0.186&0.143&0.074&\textbf{0.068}&0.112&\underline{0.072}\\
         eee\_02&2.056&0.081&0.508&0.416&0.141&\textbf{0.067}&0.075&0.092&\underline{0.071}\\
         eee\_03&0.615&0.122&0.220&0.326&0.186&0.117&0.109&\underline{0.109}&\textbf{0.087}\\
         nya\_01&0.086&0.085&0.154&0.160&0.110&0.077&\textbf{0.060}&0.086&\underline{0.075}\\
         nya\_02&0.098&0.095&0.200&0.159&0.154&0.091&0.095&\underline{0.093}&\textbf{0.083}\\
         nya\_03&0.455&0.086&0.202&0.203&0.175&0.147&0.103&\underline{0.094}&\textbf{0.083} \\
         rtp\_01&\ding{55}&0.259&2.163&0.240&0.371&0.273&\textbf{0.125}&0.326&\underline{0.129} \\
         rtp\_02&0.143&0.429&0.567&0.516&0.605&0.154&\underline{0.131}&0.137&\textbf{0.111}\\
         rtp\_03&\ding{55}&0.146&\ding{55}&0.223&0.284&\textbf{0.106}&0.137&0.143&\underline{0.121}\\
         sbs\_01&0.205&0.097&0.178&0.165&0.146&0.088&\underline{0.084}&0.097&\textbf{0.083}\\
         sbs\_02&0.528&0.097&0.165&0.176&0.133&\underline{0.089}&\textbf{0.076}&0.094&0.091\\
         sbs\_03&0.708&0.100&0.154&0.375&0.146&0.083&\textbf{0.076}&0.095&\underline{0.080}\\
         spms\_01&\ding{55}&1.125&\ding{55}&\ding{55}&1.426&0.626&\underline{0.210}&1.069&\textbf{0.162}\\
         spms\_02&\ding{55}&\ding{55}&\ding{55}&\ding{55}&1.698&\ding{55}&\underline{0.336}&\ding{55}&\textbf{0.275}\\
         spms\_03&\ding{55}&\ding{55}&\ding{55}&1.005&0.732&\ding{55}&\textbf{0.175}&\ding{55}&\underline{0.192}\\
         tnp\_01&2.969&0.320&0.879&1.937&0.159&\underline{0.081}&0.090&\textbf{0.080}&\underline{0.081}\\
         tnp\_02&0.653&0.292&1.518&\ding{55}&0.139&0.138&0.109&\underline{0.079}&\textbf{0.074}\\
         tnp\_03&0.887&0.163&1.179&\ding{55}&0.141&\ding{55}&\textbf{0.089}&\textbf{0.089}&\underline{0.101}\\
         \hline
         exp04&0.080&\ding{55}&\ding{55}&\ding{55}&\ding{55}&-&\textbf{0.022}&0.098&\underline{0.069}\\
         exp05&0.070&\ding{55}&\ding{55}&\ding{55}&\ding{55}&-&\textbf{0.020}&0.081&\underline{0.067}\\
         exp06&\ding{55}&\ding{55}&\ding{55}&\ding{55}&\ding{55}&-&\textbf{0.037}&0.190&\underline{0.133}\\
         exp14&0.120&\ding{55}&\ding{55}&\ding{55}&\ding{55}&-&\textbf{0.070}&0.888&\underline{0.092}\\
         exp16&\ding{55}&\ding{55}&\ding{55}&\ding{55}&\ding{55}&-&\underline{1.129}&\ding{55}&\textbf{0.912}\\
         exp18&\ding{55}&\ding{55}&\ding{55}&\ding{55}&\ding{55}&-&\textbf{0.231}&\ding{55}&\underline{0.236}\\
     \hline
     \hline
    \end{tabular}
\end{table*}

From Table \ref{tab:localization}, first analysing the results on VIRAL, it is evident that the successful integration of the IMU in our approach enables KN-LIO to achieve higher accuracy and better robustness compared to pure LiDAR-based dense SLAM methods (such as the neural field-based PIN-SLAM and the geometric mesh-based SLAMesh). Our KN-LIO outperforms the excellent-performing LIO-SAM in geometric feature-based methods. 
Compared to traditional direct approaches (D-LIOM, Ct-LIO, and dlio), our neural field-based pose estimation achieves higher accuracy, whether in semi-coupled or tightly coupled way. 
Compared to the state-of-the-art direct method Fast-LIO2, our performance is still comparable, with Fast-LIO2 achieving the best results on 7 sequences and our approach achieving the best on 8 sequences. 
Comparing the results of Semi-KN-LIO and KN-LIO, although semi-coupling significantly improves positioning accuracy compared to pure LiDAR, its performance is not as good as tightly coupled methods in terms of accuracy and robustness, especially on challenging sequences. 
It can be seen that although tightly coupled methods are more complex, their lossless information fusion process can better leverage the complementary characteristics of sensors.

When tested under the HILTI22 scenario, although our positioning accuracy is not as good as the state-of-the-art Fast-LIO2 on the relatively simple exp04, exp05, and exp06 sequences, on the more complex sequences exp14, exp16, and exp18, our accuracy is comparable to Fast-LIO2, with no significant difference. Furthermore, compared to traditional methods other than Fast-LIO2, we achieve significantly better robustness, enabling successful positioning and incremental reconstruction on all sequences.

\subsection{Results of Mapping}
\subsubsection{Qualitative}
\begin{figure*}[htp]
        \centering
        \includegraphics[width=0.95\textwidth]{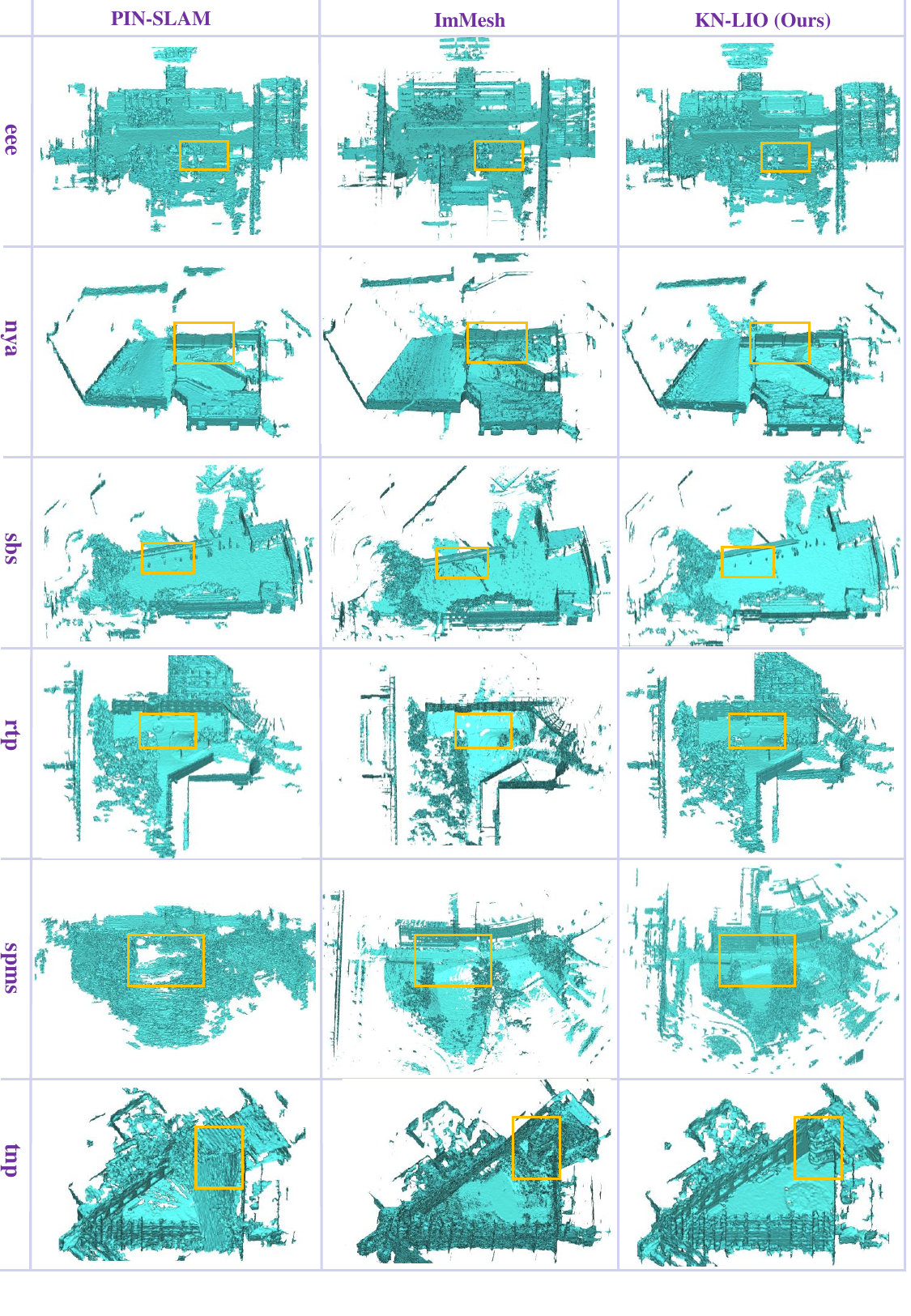}
        \caption{Meshes reconstructed on VIRAL experimental sites by PIN-SLAM, ImMesh, and our KN-LIO.}
        \label{fig-mesh}
\end{figure*}
For the 6 scenes and 18 sequences in the VIRAL dataset, we conducted separate tests using PIN-SLAM, ImMesh, and our KN-LIO. We plotted the mapping results of each algorithm achieving the best pose tracking in each scene in Fig. \ref{fig-mesh}. It can be observed that in terms of completeness, KN-LIO demonstrates the best capability, for example, in the grounds in ``rtp'' and ``spms''. In terms of detailed reconstruction, KN-LIO also exhibited higher accuracy, such as in the trees in ``eee'', the streetlights in ``sbs'', and the spiral staircase in ``tnp''. Furthermore, we found that the geometry-based ImMesh method tends to produce more noise compared to the neural field-based PIN-SLAM and KN-LIO, which is detrimental to detailed reconstruction.

\subsubsection{Quantitative}
In terms of quantitative mapping results, Table \ref{tab:mapping} presents the relevant results of different mapping algorithms at Newer College. VDB-Fusion~\cite{vdbfusion}, SHINE-Mapping~\cite{shinemapping}, and NKSR~\cite{huang2023nksr} are pure mapping algorithms, with their poses provided by the state-of-the-art LiDAR Odometry method KissICP~\cite{kissicp}, while PUMA~\cite{puma}, SLAMesh~\cite{slamesh}, NeRF-LOAM~\cite{nerfloam}, S$^2$KAN-SLAM, ImMesh~\cite{imesh}, PIN-SLAM~\cite{pinslam}, and our KN-LIO estimate poses online within their algorithms. The data in Table \ref{tab:mapping} shows that our KN-LIO performed well across all metrics. Compared to the state-of-the-art LiDAR-based mapping method PIN-SLAM, KN-LIO significantly improves reconstruction accuracy and completeness through effective integration of the IMU. In comparison to ImMesh, which estimates poses based on Fast-LIO2, the superior mapping performance of KN-LIO indicates that the learning-based approach is more conducive to the effective fusion of multi-frame observation data, leading to noise reduction and accuracy improvement.

\begin{table}[]
    \caption{Mapping stats on Newer College's `quad'.}
    \label{tab:mapping}
    \centering
    \begin{tabular}{c|c|c|c|c|c}
     \hline
     \hline
         Method & Pose & Acc. & Comp. & C-L1 & F-Score\\
         \hline
         VDB-Fusion\cite{vdbfusion} & \multirow{3}{*}{KissICP\cite{kissicp}} & 14.03 & 25.46 & 19.75 & 69.50\\
         SHINE\cite{shinemapping} & ~ & 14.87 & 20.02 & 17.45 & 68.85\\
         NKSR\cite{huang2023nksr} & ~ &15.67 & 36.87 & 26.67 & 58.57\\
         \hline
         PUMA\cite{puma} & \multirow{7}{*}{Odometry} & 15.30 & 71.91 & 43.60 & 57.27\\
         SLAMesh\cite{slamesh} & ~ & 19.21 & 48.83 & 34.02 & 45.24\\
         NeRF-LOAM\cite{nerfloam} & ~&12.89 & 22.21 & 17.55 & 74.37\\
         S$^2$KAN-SLAM\cite{s2kanslam}&~&13.32&18.80&16.06&72.03\\
         ImMesh\cite{imesh}&~& 15.05&19.80&17.42&66.87\\
         PIN-SLAM\cite{pinslam} & ~&11.55 & 15.25 & 13.40 & 82.08\\
         KN-LIO (ours)&~&\textbf{8.18}&\textbf{11.65}&\textbf{9.92}&\textbf{91.01}\\
     \hline
     \hline
    \end{tabular}
\end{table}

\subsection{Multi-LiDAR Support}
\begin{figure}[htp]
        \centering
        \includegraphics[width=0.5\textwidth]{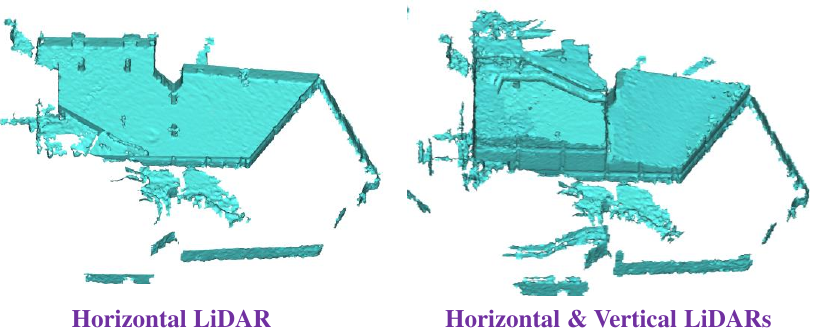}
        \caption{Mapping with the horizontal LiDAR (left) and with both the horizontal and vertical LiDARs (right).}
        \label{fig-two-lidar}
\end{figure}

To validate (Semi-)KN-LIO's support for multiple LiDAR inputs, we simultaneously utilized the horizontal and vertical LiDAR data from the VIRAL dataset to analyze their impact on the system's localization and mapping performance. As shown in Table \ref{tab:rmsetwolidar}, with the fusion of multi-view data, our approach exhibited higher pose estimation accuracy and better robustness compared to traditional probabilistic voxel-based methods. Compared to the results of single LiDAR scenarios in the table, it can be observed that the improvement in positioning accuracy due to complementary views is also significant. Furthermore, with dual LiDAR inputs, the advantages of tightly coupled fusion over semi-coupled integration are more pronounced.
Fig. \ref{fig-two-lidar} illustrates the reconstructed map under single/dual LiDAR inputs. It is evident that KN-LIO effectively supports the input of multi-view LiDAR, resulting in a more comprehensive reconstructed environmental map.

\begin{table}[htp]
    \caption{RMSEs (m) on VIRAL with dual LiDAR inputs.}
    \label{tab:rmsetwolidar}
    \centering
    \begin{tabular}{c|c|c|c|c}
     \hline
     \hline
         ~&D-LIOM-HV&Ct-LIO-HV&Semi-KN-LIO&KN-LIO \\
         \hline
         eee\_01&0.124&0.132&0.095&0.072 \\
         eee\_02&0.122&0.247&0.091& 0.067 \\
         eee\_03&0.150&0.137&0.099&0.087\\
         nya\_01&0.137&0.108&0.073&0.071 \\
         nya\_02&0.161&0.179&0.090&0.077\\
         nya\_03&0.159&0.240&0.081&0.076 \\
         rtp\_01&0.240&0.211&0.196&0.142 \\
         rtp\_02&0.192&0.223&0.156&0.108 \\
         rtp\_03&0.223&0.335&0.112&0.107 \\
         sbs\_01&0.136&0.170&0.085&0.081 \\
         sbs\_02&0.125&0.138&0.093&0.098 \\
         sbs\_03&0.112&0.152&0.067&0.089 \\
         spms\_01&\ding{55}&\ding{55}&\ding{55}&0.143 \\
         spms\_02&\ding{55}&\ding{55}&\ding{55}&0.295 \\
         spms\_03&1.005&1.015&\ding{55}&0.238 \\
         tnp\_01&0.130&0.178&0.077&0.076 \\
         tnp\_02&0.143&0.168&0.058&0.059\\
         tnp\_03&0.134&0.170&0.088&0.099\\
     \hline
     \hline
    \end{tabular}
\end{table}

\section{Conclusion}\label{sec-conclusion}
To achieve accurate simultaneous pose estimation and dense mapping, we propose a coupled LiDAR-Inertial Odometry system that integrates geometric kinematics with neural fields, (Semi-)KN-LIO. We fuse laser and inertial sensor information through an error-state Kalman filter. In the semi-coupled mode, the point cloud is first registered with the neural field, and then the error-state Kalman filter is updated with the composite registration result. In the tightly coupled mode, the system iteratively updates the error-state filter with the Kalman gain using residuals between the point cloud and the neural field until convergence. By coupling geometric kinematics with neural fields, our KN-LIO achieves comparable localization results to existing state-of-the-art LIO solutions on multiple public datasets and generates higher-quality dense mapping results than those of pure LiDAR solutions. Furthermore, support for multiple LiDARs enables our (Semi-)KN-LIO to easily fuse point cloud data from different perspectives.


%





\ifCLASSOPTIONcaptionsoff
  \newpage
\fi

\bibliographystyle{IEEEtran}
\normalem
\bibliography{IEEEabrv,ref}

\end{document}